\documentclass[letterpaper]{article} %
\usepackage{aaai18}  %
\usepackage{times}  %
\usepackage{helvet}  %
\usepackage{courier}  %
\usepackage{url}  %
\usepackage{graphicx}  %

\usepackage{latexsym}
\usepackage{amsmath}
\usepackage{amsfonts}
\usepackage{booktabs}
\usepackage[subrefformat=parens]{subcaption}

\newcommand{\Cset}{\mathbb{C}}
\newcommand{\Rset}{\mathbb{R}}
\newcommand{\mat}[1]{\boldsymbol{\mathbf{#1}}}
\newcommand{\RE}{\mathop{\text{Re}}}
\newcommand{\IM}{\mathop{\text{Im}}}
\newcommand{\sign}{\mathop{\text{sign}}}

\newcommand{\transpose}{^{\mathrm{T}}}
\newcommand{\diag}{\mathop{\text{diag}}}

\newsavebox\tempbox
\let\svwidetilde\widetilde
\newcommand\Widetilde[1]{\sbox\tempbox{$#1$}\svwidetilde{\usebox{\tempbox}}}
\newcommand{\average}[1]{\Widetilde{#1}}

\usepackage{xcolor}

\begin{document}
\title{Data-dependent Learning of Symmetric/Antisymmetric Relations\\ for Knowledge Base Completion}
\author{Hitoshi Manabe$^1$ \qquad
  Katsuhiko Hayashi$^2$ \qquad
  Masashi Shimbo$^1$ \\
  $^1$ Nara Institute of Science and Technology, \url{{manabe.hitoshi.me0, shimbo}}@is.naist.jp \\
  $^2$ NTT Communication Science Laboratories, \url{hayashi.katsuhiko@lab.ntt.co.jp}
}
\maketitle

\begin{abstract}

Embedding-based methods for knowledge base completion (KBC) learn representations of entities and relations in a vector space,
along with the scoring function to estimate the likelihood of relations between entities.
The learnable class of scoring functions is designed to be expressive enough to cover a variety of real-world relations,
but this expressive comes at the cost of an increased number of parameters.
In particular, parameters in these methods are superfluous for relations
that are either symmetric or antisymmetric.
To mitigate this problem,
we propose a new L1 regularizer for Complex Embeddings, which is one of the state-of-the-art embedding-based methods for KBC.
This regularizer promotes symmetry or antisymmetry of the scoring function on a relation-by-relation basis, in accordance with the observed data.
Our empirical evaluation shows that the proposed method outperforms the original Complex Embeddings and other baseline methods
on the FB15k dataset.

\end{abstract}

\section{Introduction}

Large-scale knowledge bases, such as YAGO \cite{yago}, Freebase \cite{freebase}, and WordNet \cite{wordnet}
are utilized in knowledge-oriented applications such as question answering and dialog systems.
Facts are stored in these knowledge bases as triplets of form
$(\text{\textit{subject entity}},\allowbreak \text{\textit{relation}},\allowbreak \text{\textit{object entity}})$.

Although a knowledge base may contain more than a million facts,
many facts are still missing \cite{survey}.
\emph{Knowledge base completion} (KBC) aims to find such missing facts automatically.
In recent years, vector embedding of knowledge bases has been actively pursued as a promising approach to KBC.
In this approach,
entities and relations %
are embedded into a vector space as their representations,
in most cases as vectors and sometimes as matrices.
A variety of methods have been proposed, each of which computes the likeliness score of given triplets
using different vector/matrix operations over the representations of entities and relations involved.

In most of the previous methods, the scoring function is designed to cover general non-symmetric relations,
i.e., relations $r$ such that for some entities $e_1$ and $e_2$, triplet $(e_1,r,e_2)$ holds but not $(e_2, r, e_1)$.
This reflects the fact that the subject and object in a relation are not interchangeable in general (e.g., \emph{parent\_of}).

However, the degree of symmetry differs from relation to relation.
In particular,
a non-negligible number of symmetric relations exist in knowledge bases (e.g., \emph{sibling\_of}).
Moreover,
many non-symmetric relations in knowledge base are actually \emph{antisymmetric},
in the sense that for \emph{every} distinct pair $e_1, e_2$ of entities, if $(e_1,r,e_2)$ holds, then $(e_2, r, e_1)$ never holds.%
\footnote{Note that even if a relation is antisymmetric in the above sense, its truth-value matrix may not be antisymmetric.
}
For relations that show certain regularities such as above, the expressiveness of models to capture general relations might be superfluous,
and a model with less parameters might be preferable.

It thus seems desirable to encourage the scoring function to produce sparser models,
if the observed data suggests a relation being symmetric or antisymmetric.
As we do not assume any background knowledge about individual relations,
the choice between these contrasting properties must be made solely from the observed data.
Further, we do not want the model to sacrifice the expressiveness to cope with relations that are neither purely symmetric or antisymmetric.

Complex Embeddings (ComplEx) \cite{complex} are one of the state-of-the-art methods for KBC.
ComplEx represents entities and relations as complex vectors,
and it can model general non-symmetric relations thanks to the scoring function defined by the Hermitian inner product of these vectors.
However, for symmetric relations, the imaginary parts in relation vectors are redundant parameters since they only contribute to non-symmetry of the scoring function.
ComplEx is thus not exempt from the issue we mentioned above: the lack of symmetry/antisymmetry consideration for individual relations.
Our experimental results show that this issue indeed impairs the performance of ComplEx.

In this paper, we propose a technique for training ComplEx
relation vectors adaptively to the degree of symmetry/antisymmetry
observed in the data.
Our method is based on L1 regularization, but not in the standard way;
the goal here is not to make a sparse, succinct model but to adjust the degree of symmetry/antisymmetry on the relation-by-relation basis,
in a data-driven fashion.
In our model, L1 regularization is imposed on the products of \emph{coupled} parameters,
with each parameter contributing to either the symmetry or antisymmetry of the learned scoring function.

Experiments with synthetic data show that
our method works as expected:
Compared with the standard L1 regularization,
the learned functions is more symmetric for symmetric relations and more antisymmetric for antisymmetric relations.
Moreover, in KBC tasks on real datasets,
our method outperforms the original ComplEx with standard L1 and L2 regularization, as well as other baseline methods.

\section{Background}

Let %
$\Rset$ be the set of reals,
and $\Cset$ be the set of complex numbers.
Let $i \in \Cset$ denote the imaginary unit.
$[\mat{v}]_j$ denotes the $j$th component of vector $\mat{v}$,
and $[\mat{M}]_{jk}$ denotes the $(j,k)$-element of matrix $\mat{M}$.
A superscript \text{T} (e.g., $\mat{v}\transpose$) represents vector/matrix transpose.
For a complex scalar, vector, or matrix $\mat{Z}$,
$\overline{\mat{Z}}$ represent its complex conjugate,
with $\RE(\mat{Z})$ and $\IM(\mat{Z})$ denoting its real and imaginary parts, respectively.

\subsection{Knowledge Base Completion}

Let $\mathcal{E}$ and $\mathcal{R}$ respectively be the sets of entities and the (names of) binary relations over entities in an incomplete knowledge base.
Suppose a relational triplet $(s,r,o)$ is not in the knowledge base for some $s, o \in \mathcal{E}$ and $r \in \mathcal{R}$.
The task of KBC is to determine the truth value of such an unknown triplet;
i.e., whether relation $r$ holds between subject entity $s$ and object entity $o$.

A typical approach to KBC is to learn a \emph{scoring function} $\phi(s, r, o)$
to estimate the likeliness of an unknown triplet $(s, r, o)$,
using as training data the existing triplets in the knowledge base and their truth values.
A higher score indicates that the triplet is more likely to hold.

The scoring function $\phi$ is usually parameterized,
and the task of learning $\phi$ is recast as
that of tuning the model parameters. %
To indicate this explicitly, model parameters $\mat{\Theta}$ are sometimes included in the arguments of the scoring function,
as in $\phi(s, r, o; \mat{\Theta})$.

\subsection{Complex Embeddings (ComplEx)}

The embedding-based approach to KBC defines the scoring function in terms of the vector representation (or, \emph{embeddings}) of entities and relations.
In this approach, model parameters $\mat{\Theta}$ consist of these representation vectors.

ComplEx \cite{complex} is one of the latest embedding-based methods for KBC.
It represents entities and relations as complex vectors.
Let $\mat{e}_j, \mat{w}_r \in \mathbb{C}^d$ respectively denote the $d$-dimensional complex vector representations of
entity $j \in \mathcal{E}$ and relation $r \in \mathcal{R}$.
The scoring function of ComplEx %
is defined by
\begin{align}
  \phi(s, r, o; \mat{\Theta}) & = \RE \left( \mat{e}_s\transpose \diag(\mat{w}_r) \overline{\mat{e}_o} \right) \label{eq:complex-score-vector-form}  \\
                              & = \RE \left( \langle \mat{w}_r, \mat{e}_s, \overline{\mat{e}_o} \rangle \right) \nonumber           \\
                              & =  \langle \RE (\mat{w}_r), \RE (\mat{e}_s), \RE (\mat{e}_o) \rangle \nonumber                      \\
                              & \qquad + \langle \RE (\mat{w}_r), \IM (\mat{e}_s), \IM (\mat{e}_o) \rangle \nonumber                 \\
                              & \qquad + \langle \IM (\mat{w}_r), \RE (\mat{e}_s), \IM (\mat{e}_o) \rangle \nonumber                 \\
                              & \qquad - \langle \IM (\mat{w}_r), \IM (\mat{e}_s), \RE (\mat{e}_o) \rangle, \label{eq:complex-score-breakdown}
\end{align}
where $\diag(\mat{v})$ denotes a diagonal matrix with the diagonal given by vector $\mat{v}$,
and
$\langle \mat{u}, \mat{v}, \mat{w}\rangle = \left( \sum_{k=1}^d [\mat{u}]_k [\mat{v}]_k [\mat{w}]_k \right) $,
with $\mat{\Theta} = \{ \mat{e}_j \in \Cset^d \mid j \in \mathcal{E} \} \cup \{ \mat{w}_r \in \Cset^d \mid r \in \mathcal{R} \}$.
The use of complex vectors and Hermitian inner product
makes ComplEx both expressive and computationally efficient.

\section{Learning (Anti)symmetric Relations with L1 Regularization}

\subsection{Roles of Real/Imaginary Parts in Relation Vectors}

Many relations in knowledge bases are either symmetric or antisymmetric.
For example, all 18 relations in WordNet
are either symmetric (4 relations) or antisymmetric (14 relations).
Also, relations that take different ``types'' of entities as the subject and object
are necessarily antisymmetric;
take relation \textit{born\_in} for example, which is defined for a person and a location.
Clearly, if
$(\textit{Barack\_Obama}, \textit{born\_in}, \textit{Hawaii})$ holds, then $( \textit{Hawaii}, \textit{born\_in}, \textit{Barack\_Obama})$ does not.

Now, let us look closely at the scoring function of ComplEx given by Eq.~\eqref{eq:complex-score-vector-form}.
We observe the following:
If the relation vector $\mat{w}_r$ is a real vector, then
$\phi(s,r,o) = \phi(o,r,s)$ for any $s, o\in \mathcal{E}$;
i.e., the scoring function $\phi(s,r,o)$ is symmetric with respect to $s$ and $o$.
This can be seen by substituting $\IM(\mat{w}_r) = \mat{0}$ in Eq.~\eqref{eq:complex-score-breakdown}, in which case the last two terms vanish.
If, to the contrary,
$\mat{w}_r$ is purely imaginary, %
$\phi(s,r,o)$ is antisymmetric in $s$ and $o$, in the sense that $\phi(s,r,o) = -\phi(o,r,s)$.
Again, this can be verified with Eq.~\eqref{eq:complex-score-breakdown}, but this time by substituting $\RE(\mat{w}_r) = \mat{0}$.

As we see from these two cases,
the real parts in the components of $\mat{w}_r$ are responsible for making the scoring function $\phi$ symmetric,
whereas the imaginary parts in $\mat{w}_r$ are responsible for making it antisymmetric.

Each relation has a different degree of symmetry/\linebreak anti\-symmetry,
but the original ComplEx, which is usually trained with L2 regularization, does not take this difference into account.
Specifically, 
L2 regularization is equivalent to making a prior assumption that all parameters,
including the real and imaginary parts of relation vectors,
are independent. %
As we have discussed above,
this independence assumption is unsuited for symmetric and antisymmetric relations.
For instance, we expect the vector for symmetric relations to have a large number of nonzero real parts and zero imaginary parts.

\subsection{Multiplicative L1 Regularization for Coupled Parameters}

On the basis of the observation above, we introduce a new regularization term for training ComplEx vectors.
This term encourages individual relation vectors to be more symmetric or antisymmetric in accordance with the observed data.
The resulting objective function is
\begin{align}
  \underset{\mat{\Theta}}{\mathrm{min}} \sum_{(s, r, o) \in \Omega} \log(1 + \exp(-y_{rso} \phi(s, r, o; \mat{\Theta}))) \nonumber \label{eq:objective} \\
  + \lambda ( \alpha R_1(\mat{\Theta}) + (1 - \alpha) R_2(\mat{\Theta}))
\end{align}
where $\Omega$ is the training samples of triplets;
$y_{rso} \in \{+1, -1\}$ gives the truth value of the triplet $(s, r, o)$;
hyperparameter $\lambda \geq 0$ determines the overall weight on the regularization terms;
and $\alpha \in [0, 1]$ (also a hyperparameter) controls the balance between
two regularization terms $R_1$ and $R_2$.
These terms are defined as follows:
\begin{align}
  R_1(\mat{\Theta}) %
                    & = \sum_{r \in \mathcal{R}} \sum_{k=1}^d \left| \RE ([\mat{w}_r]_k) \cdot \IM ([\mat{w}_r]_k) \right|,  \label{eq:r1} \\
  R_2(\mat{\Theta}) & = \left\| \mat{\Theta} \right\|^2_2. \label{eq:r2}
\end{align}
In Eq.~\eqref{eq:r2}, $\mat{\Theta}$ is treated as a vector, with all the parameters it contains as the vector components.

Eq.~\eqref{eq:objective} differs from the original ComplEx objective in that it introduces the proposed regularizer $R_1$, %
which is a form of L1-norm penalty (see Equation~\eqref{eq:r1}).
In general, L1-norm penalty terms promote producing sparse solutions for the model parameters
and are used for feature selection or the improve the interpretability of the model.
Note, however, that our L1 penalty encourages sparsity of \emph{pairwise} products.
This means that only one of the coupled parameters needs to be propelled towards zero to minimize the summarized in Eq.~\eqref{eq:r1}.
To distinguish from the standard L1 regularization, we call the regularization term in $R_1$ \emph{multiplicative L1 regularizer},
since it is based on the L1 norm of the vector of the product of the real and imaginary parts of each component.

As we explain in the next subsection,
standard L1 regularization implies the independence of parameters as a prior.
By contrast, our regularization term $R_1$ dictates the interaction between the real and imaginary parts of a component in a relation vector;
if, as a result of L1 regularization, one of these parts falls to zero, the other can freely move to minimize the objective~\eqref{eq:objective}.
This encourages selecting either of the coupled parameters to be zero, but not necessarily both.

Unlike the standard L1 regularization, the proposed regularization term is non-convex, and makes the optimization harder\footnote{
  Notice that the objective function in ComplEx is already non-convex without a regularization term.}.
However,
in our experiments reported below, multiplicative L1 regularization outperforms the standard one in KBC, and is robust against random initialization.

Since the real and imaginary parts of a relation vector govern the symmetry/antisymmetry of the scoring function for the relation,
this L1 penalty term is expected to help guide learning a vector for relation $r$ in accordance with whether $r$ is symmetric, antisymmetric, or neither of them,
as observed in the training data.
For example, if the data suggests $r$ is likely to be symmetric,
our L1 regularizer should encourage the imaginary parts to be zero while allowing the real parts to take on arbitrary values.
Because parameters are coupled componentwise,
the proposed model can also cope with non-symmetric, non-antisymmetric relations with different degree of symmetry/antisymmetry.

\subsection{MAP Interpretation}

MAP estimation finds the best model parameters $\hat{\mat{\Theta}}$ by maximizing a posterior distribution:
\begin{align}
  \hat{\mat{\Theta}} & = \underset{\mat{\Theta}}{\rm argmax} \log p(\mat{\Theta} | \mathcal{D}) \nonumber \\
  & = \underset{\mat{\Theta}}{\rm argmax} \log p(\mathcal{D} | \mat{\Theta}) + \log p(\mat{\Theta}), \label{eq:map}
\end{align}
where $\mathcal{D}$ is the observed data.
The first term %
represents the likelihood function, and the second term represents the prior distribution of parameters.

Our objective function Eq.~\eqref{eq:objective} can also be viewed as MAP estimation in the form of Eq.~\eqref{eq:map};
the first term in our objective corresponds to the likelihood function,
and the regularizer terms define the prior.

Let us discuss the prior distribution implicitly assumed by using the proposed multiplicative L1 regularizer $R_1$.%
\footnote{
  For brevity, we neglect the regularizer $R_2$ and focus on $R_1$ in this discussion.
  }
Let $C, C', C'', \ldots$ denote constants.
Our multiplicative L1 regularization is equivalent to assuming the prior %
\begin{align*}
  p(\mat{\Theta})                        & =    \prod_{r\in \mathcal{R}}  p(\mat{w}_r) \\ %
  \intertext{with}
  p(\mat{w}_r)                           & = \prod_{k=1}^d C \exp \left( - \frac{ \left|\RE ([\mat{w}_r]_k) \cdot \IM ([\mat{w}_r]_k) \right| }{C'} \right).
\end{align*}
In other words, a $0$-mean Laplacian prior is assumed on the distribution of $\RE ([\mat{w}_r]_k) \cdot \IM ([\mat{w}_r]_k)$.
The equivalence can be seen by
\begin{align*}
  \log p(\mat{\Theta}) & = \log \prod_{r\in \mathcal{R}} p( \mat{w}_r) \nonumber                                                                                                            \\
                                     & = \log \! \prod_{r\in \mathcal{R}} \prod_{k=1}^d \! C \exp \! \left( \!\! - \frac{ \left|\RE ([\mat{w}_r]_k) \cdot \IM ([\mat{w}_r]_k) \right| }{C'} \right) \nonumber \\
                                     & = - C'' \! \sum_{r\in \mathcal{R}} \sum_{k=1}^d \left|\RE ([\mat{w}_r]_k) \cdot \IM ([\mat{w}_r]_k) \right| + C'''.
\end{align*}
Neglecting the scaling factor $C''$ and the constant term $C'''$, we see that this is equal to the regularizer $R_1$ in Eq.~\eqref{eq:r1}.

Now,
suppose the standard L1 regularizer
\begin{equation}
  R_{\text{std L1}}(\mat{\Theta}) = \sum_{r\in\mathcal{R}} \sum_{k=1}^d ( \left| \RE ([\mat{w}_r]_k) \right| + \left| \IM ([\mat{w}_r]_k) \right|)
  \label{eq:std-l1-term}
\end{equation}
is instead of the proposed $R_1(\Theta)$.
In terms of MAP estimation, in this case, its use is equivalent to assuming a prior distribution
\begin{equation}
  p(\mat{w}_r) = \prod_{k=1}^d  p(\RE ([\mat{w}_r]_k)) \, p(\IM ([\mat{w}_r]_k))
  \label{eq:std-l1-prior}
\end{equation}
where both $p(\RE ([\mat{w}_r]_k))$ and $p(\IM ([\mat{w}_r]_k))$ obey a 0-mean Laplacian distribution.
Notice that the distribution \eqref{eq:std-l1-prior} assumes the independence of the real and imaginary parts of components.
This assumption can be harmful if parameters are (positively or negatively) correlated with each other,
which is indeed the case with symmetric or antisymmetric relations.

\subsection{Training Procedures}

Optimizing our objective function (Eq.~\eqref{eq:objective}) is difficult with standard online optimization methods, such as stochastic gradient descent.
In this paper, we extend the Regularized Dual Averaging (RDA) algorithm \cite{rda}, which can produce sparse solutions effectively and is used in learning sparse word representations \cite{overcomplete,Sun:Sparse}.
Let us indicate the parameter value at time $t$ by superscript $(t)$.
RDA keeps track of the online average subgradients at time $t$: $\average{\mat{g}}^{(t)} = (1/t) \sum_{\tau=1}^t \mat{g}^{(\tau)}$,
where $\mat{g}^{(t)}$ is the subgradient at time $t$.
In this paper, we calculate the subgradients $\mat{g}^{(t)}$ in terms of only the loss function and L2 norm penalty term;
i.e., they are the derivatives of the objective without the L1 penalty term.

The update formulas for multiplicative L1 regularizer are given as follows:
\begin{align*}
  \RE ([\mat{w}_r]_k^{(t+1)})                 & = { \begin{cases}
                                      0,      & \text{if } \left| [ \average{\mat{g}_r}]_k^{(t)} \right| \leq \beta \left| \IM ([\mat{w}_r]_k^{(t)}) \right|,   \\
                                      \gamma, & \text{otherwise,}
                                     \end{cases}
                                  }                                                                                                                             \\
  \IM ([\mat{w}_r]_k^{(t+1)})                 & = { \begin{cases}
                                      0,      & \text{if } \left| [ \average{\mat{g'}_r} ]_k^{(t)} \right| \leq \beta \left| \RE ([\mat{w}_r]_k^{(t)}) \right|, \\
                                      \gamma' & \text{otherwise,}
                                    \end{cases}
                                  }
\end{align*}
where $\beta = \lambda\alpha$ is a constant,
$\mat{g}_r, \mat{g}'_r\in \Rset^d$ are the real and imaginary parts of the subgradients with respect to relation $r$, and
\begin{align*}
  \gamma  & =  - \eta t \left( [\average{\mat{g}_r } ]_k^{(t)} - \beta \left| \IM ([\mat{w}_r]_k^{(t)}) \right| \sign([\average{\mat{g} _r} ]_k^{(t)} ) \right), \\
  \gamma' & =  - \eta t \left( [\average{\mat{g}'_r}]_k^{(t)}  - \beta \left| \RE ([\mat{w}_r]_k^{(t)}) \right| \sign([\average{\mat{g}'_r} ]_k^{(t)} )\right).
\end{align*}
From these formulas, we notice the interaction of the real and imaginary parts of a component;
the imaginary part appears in the update formula for the real part, and vice versa.
The term
$\beta |\IM ([\mat{w}_r^{(t)}]_k)|$ can be regarded as the strength of L1 regularizer specialized for $\RE ([\mat{w}_r]_k)$ at time $t$;
if $\IM ([\mat{w}_r]_k)=0$, then $\RE ([\mat{w}_r]_k)$ keeps a nonzero value $\gamma$. %
Likewise, if $\RE ([\mat{w}_r]_k)=0$, $\IM ([\mat{w}_r]_k)$ is free to take on a nonzero value. %
Notice that the above update formulas are applied only to relation vectors.
Entity vectors are learned with a standard optimization method.

\section{Related Work}
\subsection{Knowledge Base Embedding}

RESCAL~\cite{rescal} is an embedding-based KBC method whose scoring function is formulated as $\mat{e}_{s}^{\rm T}\mat{W}_{r}\mat{e}_{o}$,
where $\mat{e}_s, \mat{e}_o \in \Rset^d$ are the vector representations of entities $s$ and $o$, respectively,
and (possibly non-symmetric) matrix $\mat{W}_r \in \Rset^{d\times d}$ represents a relation $r$.
Although RESCAL is able to output non-symmetric scoring functions,
each relation vector $\mat{W}_r$ holds $d^2$ parameters.
This can be problematic both in terms of overfitting and computational cost.
To avoid this problem, several methods have been proposed recently.

DistMult~\cite{distmult} restricts the relation matrix to be diagonal, $\mat{W}_r = \diag(\mat{w}_r)$,
and it can compute the likelihood score in time $O(d)$ by way of %
$\phi(s,r,o) = \mat{e}_s\transpose \diag(\mat{w}_r) \mat{e}_o $. %
However, this form of function is necessarily symmetric in $s$ and $o$; i.e., $\phi(s,r,o)=\phi(o,r,s)$.
To reconcile efficiency and expressiveness,
\citeauthor{complex} \shortcite{complex} proposed ComplEx, %
using the complex-valued representations and Hermitian inner product to define the scoring function (Eq.~\eqref{eq:complex-score-vector-form}).
ComplEx is founded on the unitary diagonalization of normal matrices \cite{unitarydiag}.
Unlike DistMult, the scoring function can be nonsymmetric in $s$ and $o$.
\citeauthor{eq} \shortcite{eq} found that ComplEx is equivalent to another state-of-the-art KBC method, Holographic Embeddings~(HolE)~\cite{hole}.

ANALOGY~\cite{analogy} also assumes that $\mat{W}_r$ is real normal.
They showed that any real normal matrix can be block-diagonalized, where each diagonal block is either a real scalar or a $2\times 2$ real matrix
of form $\begin{pmatrix} a & b \\ -b & a \end{pmatrix}$.
Notice that this $2\times2$ matrix is exactly the real-valued encoding of a complex value $a+i b$.
In this sense, ANALOGY can be regarded as a hybrid of ComplEx (corresponding to the $2\times2$ diagonal blocks) and DistMult (real scalar diagonal elements).
It can also be viewed as an attempt to reduce of the number of parameters in ComplEx, by
constraining some of the imaginary parts of the vector components to be zero. %
Although the idea resembles our proposed method,
in ANALOGY, the reduced parameters (the number of scalar diagonal components) is a hyperparameter.
By contrast, our approach lets the data adjust the number of parameters for individual relations, by means of the multiplicative L1 regularizer.

\begin{figure*}[tbp]

\centering
\begin{tabular}{p{0.47\linewidth}p{0.47\linewidth}}
  \includegraphics[width=\linewidth]{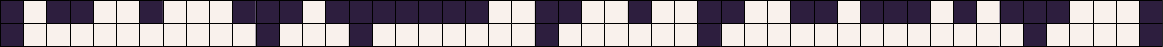}\par
  \subcaption{Symmetric relation using standard L1 regularization}
  &
    \includegraphics[width=\linewidth]{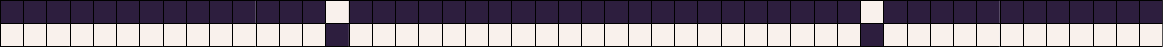}\par
    \subcaption{Symmetric relation using multiplicative L1 regularization}
  \\
  \includegraphics[width=\linewidth]{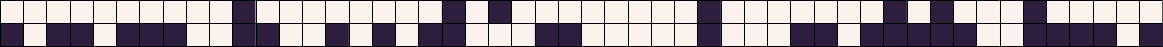}\par
  \subcaption{Antisymmetric relation using standard L1 regularization}
  &
    \includegraphics[width=\linewidth]{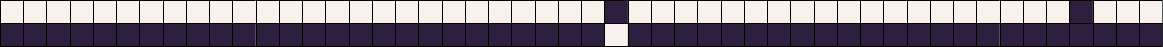}
    \subcaption{Antisymmetric relation using multiplicative L1 regularization}
  \\
  \includegraphics[width=\linewidth]{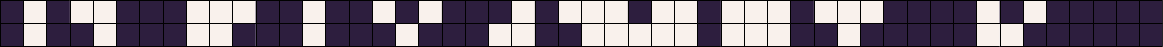}
  \subcaption{Non-symmetric/non-antisymmetric relation using standard L1 regularization}
  &
    \includegraphics[width=\linewidth]{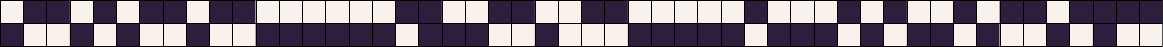}
    \subcaption{Non-symmetric/non-antisymmetric relation using multiplicative L1 regularization}
\end{tabular}

\caption{Visualization of the vector representations trained on synthetic data.
  Each column represents a complex-valued vector component,
  with the upper and lower cells representing the real and imaginary parts, respectively.
The black cells represent non-zero values.}
\label{fig:sparse}

\end{figure*}

\subsection{Sparse Modeling}

Sparse modeling is often used to improve model interpretability or to enhance performance
when the size of training data is insufficient relative to the number of model parameters.
Lasso~\cite{lasso} is a sparse modeling technique for linear regression.
Lasso uses L1 norm penalty to effectively find which parameters can be dispensed with.
Following Lasso, several variants for inducing structural sparsity have been proposed, e.g., \cite{fused,exclusivelasso}.
One of them, the exclusive group Lasso~\cite{exclusivelasso} uses an $l_{1, 2}$-norm penalty,
to enforce sparsity at an intra-group level.
An interesting connection exists between our regularization terms in Eq.~\eqref{eq:objective} and the exclusive group Lasso.
Let $R_j(\mat{w}_r)$, $j=1,2$ represents the terms in $R_j(\mat{\Theta})$ that are concerned with relation $r$.
When $\alpha = 2/3$, we can show that the regularizer terms
$\alpha R_1(\mat{w}_r) + (1 - \alpha) R_2 (\mat{w}_r) = (1/3) \sum_{k=1}^d (  \RE ([\mat{w}_r]_k) + \IM ([\mat{w}_r]_k) )^2$.
The right-hand side can be seen as an instance of the exclusive group Lasso.

Word representation learning~\cite{word2vec,glove}
has proven useful for a variety of natural language processing tasks.
Sparse modeling has been applied to improve the interpretability of the learned word vectors
while maintaining the expressive power of the model~\cite{overcomplete,Sun:Sparse}.
Unlike our proposed method, however, the improvement of the model performance was not the main focus. %

\section{Experiments}
\subsection{Demonstrations on Synthetic Data}

\begin{table}[t]
\caption{Classification accuracy (\%) on synthetic data.}
\label{tab:syn-results}
\centering
\small
\begin{tabular}{ccccc}
\toprule
Models              & sym        & anti       & other      & all        \\
\cmidrule(lr){1-1}\cmidrule(lr){2-4}\cmidrule(lr){5-5}
ComplEx w/ std L1 & 89.8       & 92.1       & {\bf 65.7} & 81.5       \\
ComplEx w/ mul L1 & {\bf 93.5} & {\bf 94.4} & 65.3       & {\bf 83.3} \\
\bottomrule
\end{tabular}
\end{table}
We conducted experiments with synthetic data to verify that
our proposed method can learn symmetric and antisymmetric relations,
given such data.

We randomly created a knowledge base of $3$ relations and $50$ entities.
We generated a total of 6,712 triplets,
and sampled 5369 triplets as training set, then one-half of the remaining triplet as validation set and the other as test set.
The truth values of triplets were determined such that
the first relation was symmetric, the second was antisymmetric, and
the last was neither symmetric or antisymmetric.

We compared the standard and multiplicative L1 regularizers on this dataset.
For the standard L1 regularization, the regularizer $R_{\text{std}}$ (Eq.~\eqref{eq:std-l1-term}) was used
in place of $R_1$ (Eq.~\eqref{eq:r1}) in the objective function \eqref{eq:objective}.
The dimension of the embedding space was set to $d=50$.
For the multiplicative L1 regularizer, hyperparameters were set as follows: $\alpha = 1.0, \lambda = 0.05, \eta = 0.1$.

Figure~\ref{fig:sparse} displays which real and imaginary parts of the learned relation vectors have non-zero values.
The multiplicative L1 regularizer produced the expected results:
most of the imaginary parts were zero in the symmetric relation and the real parts were zero in the antisymmetric relation.

Table~\ref{tab:syn-results} shows the triplet classification accuracy on the test set.
Triplet classification is the task of predicting the truth value of given triplets in the test set.
For a given triplet, the prediction of the systems was determined by the sign of the output score ($+$ = true, $-$ = false).
The multiplicative L1 regularizer (`ComplEx w/ mul L1') outperformed the standard L1 regularizer (`ComplEx w/ std L1')
considerably for both symmetric and antisymmetric relations.

\subsection{Real Datasets: WN18 and FB15k}

Following previous work, we used the WordNet (WN18) and Freebase (FB15k) datasets
to verify the benefits of our proposed method.
The dataset statistics are shown in Table~\ref{tab:stats}.
Because the datasets contain only positive triplets, (pseudo-)negative samples must be generated in this experiment.
In this experiment, negative samples were generated by replacing the subject $s$ and object $o$ in a positive triplet $(s,r,o)$
with a randomly sampled entity from $\mathcal{E}$.

\begin{table}[tb]
  \caption{Dataset statistics for FB15k and WN18.}
  \label{tab:stats}
  \small
  \centering
  \begin{tabular}{lrrrrr}
    \toprule
    Dataset & $|\mathcal{E}|$ & $|\mathcal{R}|$ & \#train & \#valid & \#test \\
    \cmidrule(lr){1-6}
    FB15k & 14,951 & 1,345 & 483,142 & 50,000 & 59,071 \\
    WN18 & 40,943 & 18 & 141,442 & 5,000 & 5,000 \\
    \bottomrule
  \end{tabular}
\end{table}

\begin{table*}[tb]
  \caption{Results on the WN18 and FB15k datasets: (Filtered and raw) MRR and filtered Hits@$\{1,3,10\}$ (\%).
    * and ** denote the results reported in~\cite{complex} and \cite{analogy}, respectively.}
  \label{tab:res}
  \small
  \begin{center}
    \begin{tabular}{lrrrrrrrrrr}
      \toprule
                               & \multicolumn{5}{c}{WN18}   & \multicolumn{5}{c}{FB15k}                                                                                                                      \\
      \cmidrule(lr){2-6}\cmidrule(lr){7-11}
                               & \multicolumn{2}{c}{MRR}    & \multicolumn{3}{c}{Hits@} & \multicolumn{2}{c}{MRR} & \multicolumn{3}{c}{Hits@}                                                                \\
      \cmidrule(lr){2-3}\cmidrule(lr){4-6}\cmidrule(lr){7-8}\cmidrule(lr){9-11}
      Models                   & Filter                     & Raw                       & 1                       & 3          & 10         & Filter     & Raw        & 1          & 3          & 10         \\
      \cmidrule(lr){1-11}
      TransE*                  & 45.4                       & 33.5                      & 8.9                     & 82.3       & 93.4       & 38.0       & 22.1       & 23.1       & 47.2       & 64.1       \\
      DistMult*                & 82.2                       & 53.2                      & 72.8                    & 91.4       & 93.6       & 65.4       & 24.2       & 54.6       & 73.3       & 82.4       \\
      HolE*                    & 93.8                       & 61.6                      & 93.0                    & 94.5       & 94.9       & 52.4       & 23.2       & 40.2       & 61.3       & 73.9       \\
      ComplEx*                 & 94.1                       & 58.7                      & 93.6                    & 94.5       & 94.7       & 69.2       & 24.2       & 59.9       & 75.9       & 84.0       \\
      ANALOGY**                & 94.2                       & {\bf 65.7}                & 93.9                    & 94.4       & 94.7       & 72.5       & 25.3       & {\bf 64.6} & 78.5       & 85.4       \\
      \cmidrule(lr){1-11}
      ComplEx ($\alpha = 0$)   & {\bf 94.3}                 & 58.2                      & {\bf 94.0}              & {\bf 94.6} & 94.8       & 69.5       & 24.2       & 59.8       & 76.9       & 85.0       \\
      ComplEx w/ std L1      & {\bf 94.3}                 & 57.9                      & {\bf 94.0}              & 94.5       & 94.8       & 71.1       & 25.5       & 61.8       & 78.3       & 85.6       \\
      ComplEx w/ mul L1      & {\bf 94.3}                 & 58.5                      & {\bf 94.0}              & {\bf 94.6} & {\bf 94.9} & {\bf 73.3} & {\bf 25.8} & 64.3       & {\bf 80.3} & {\bf 86.8} \\
      \bottomrule
    \end{tabular}
  \end{center}
\end{table*}

For evaluation, we performed the entity prediction task.
In this task, an incomplete triplet is given, which is generated by hiding one of the entities, either $s$ or $o$, from a positive triplet.
The system must output the rankings of entities in $\mathcal{E}$ for the missing $s$ or $o$ in the triplet,
with the goal of placing (unknown) true $s$ or $o$ higher in the rankings.
Systems that learn a scoring function $\phi(s,r,o)$ can use the score for computing the rankings.
The quality of the output rankings is measured by
two standard evaluation measures for the KBC task:
Mean Reciprocal Rank~(MRR) and Hits@$1$, $3$ and $10$.
We here report results in both the filtered and raw settings \cite{transe} for MRR,
but only filtered values for Hits@$n$.

\subsection{Experimental Setup}

We selected the hyperparameters $\lambda$, $\alpha$, and $\eta$ via grid search such that they maximize the filtered MRR on the validation set.
The ranges for the grid search were as follows:
$\lambda \in \{ 0.01, 0.001, 0.0001, 0 \}$,
$\alpha \in \{ 0, 0.3, 0.5, 0.7, 1.0 \}$,
$\eta \in \{ 0.1, 0.05 \}$.
During the training, learning rate $\eta$ was tuned with AdaGrad~\cite{adagrad}, both for entity and relation vectors.
The maximum number of training epochs was set to 500 and the dimension of the vector space was $d=200$.
The number of negative triplets generated per positive training triplet was 10 for FB15k
and 5 for WN18.

\begin{table}[tb]
  \caption{Results on WN18 when the size of training data is reduced to a half}
  \label{tab:res-low}
  \small
  \centering
    \begin{tabular}{lrrrrr}
      \toprule
                             & \multicolumn{2}{c}{MRR} & \multicolumn{3}{c}{Hits@}                         \\
      \cmidrule(lr){2-3}\cmidrule(lr){4-6}
      Models                 & Filter                  & Raw        & 1          & 3          & 10         \\
      \cmidrule(lr){1-6}
      ComplEx ($\alpha = 0$) & 48.3                    & 32.9       & 47.4       & 48.7       & 49.8       \\
      ComplEx w/ std L1    & 48.2                    & 33.4       & 47.2       & 48.7       & 50.2       \\
      ComplEx w/ mul L1    & {\bf 49.0}              & {\bf 34.6} & {\bf 47.7} & {\bf 49.7} & {\bf 51.2} \\
      \bottomrule
    \end{tabular}
\end{table}

\begin{figure}[t]
\begin{center}
\includegraphics[width=\linewidth]{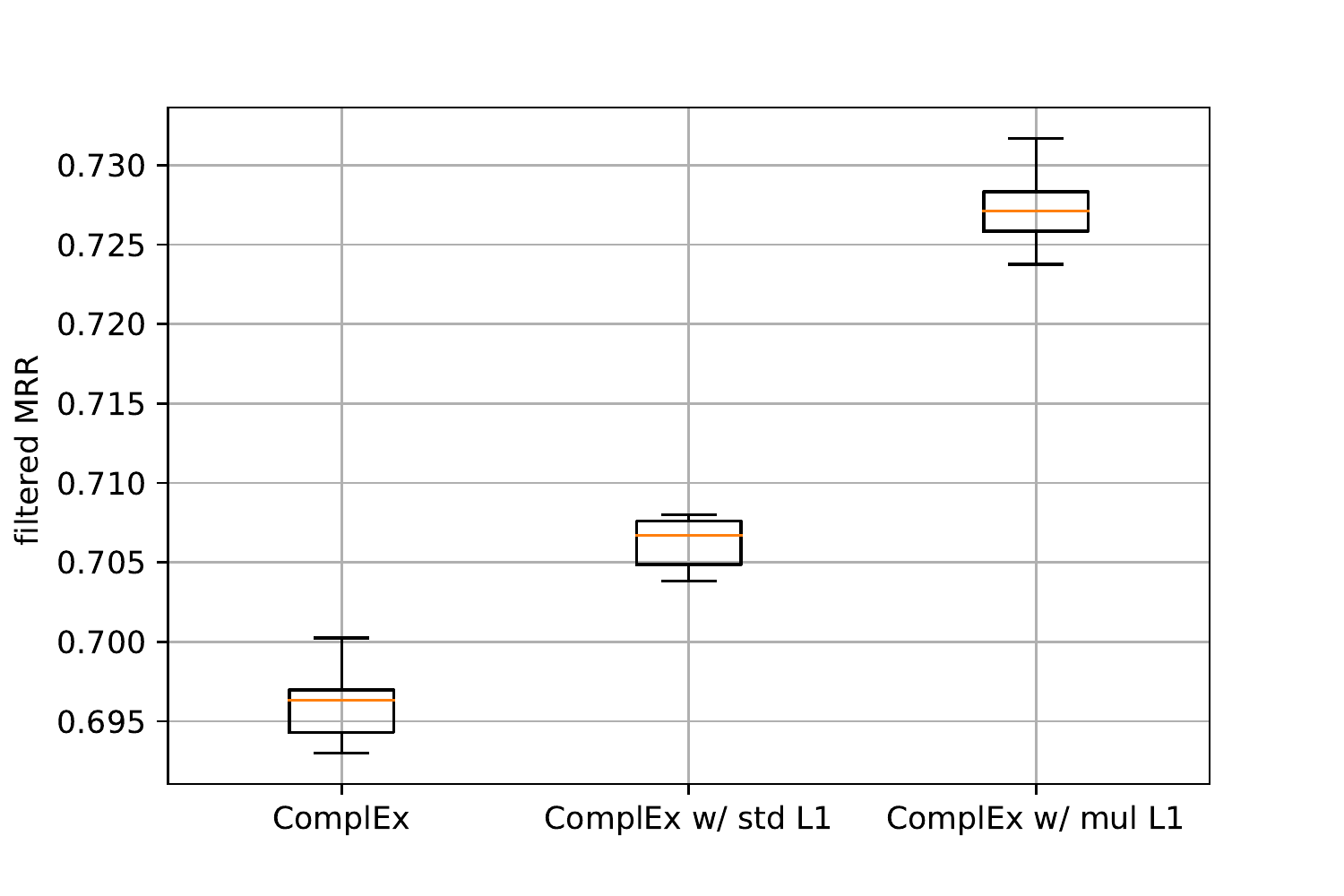}
\caption{The box plots for the variance of filtered MRR on the FB15k dataset.}
\label{fig:boxplot}
\end{center}
\end{figure}

\begin{figure*}[t]
\centering
\begin{tabular}{cc}
\begin{minipage}{0.48\hsize}
\centering
\includegraphics[width=\linewidth]{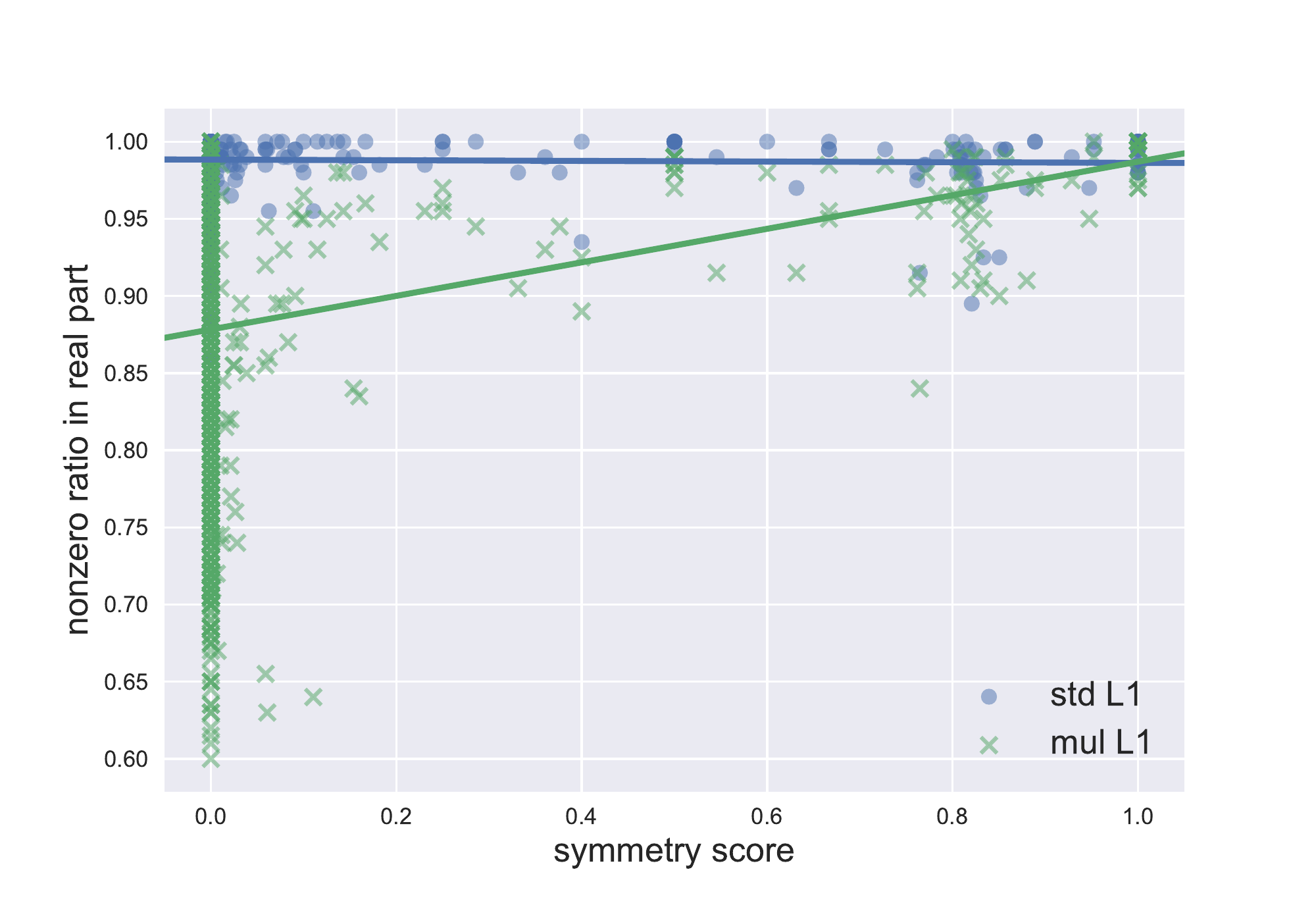}
\subcaption{The real part on the FB15k dataset}
\end{minipage}
                             &
\begin{minipage}{0.48\hsize}
\centering
\includegraphics[width=\linewidth]{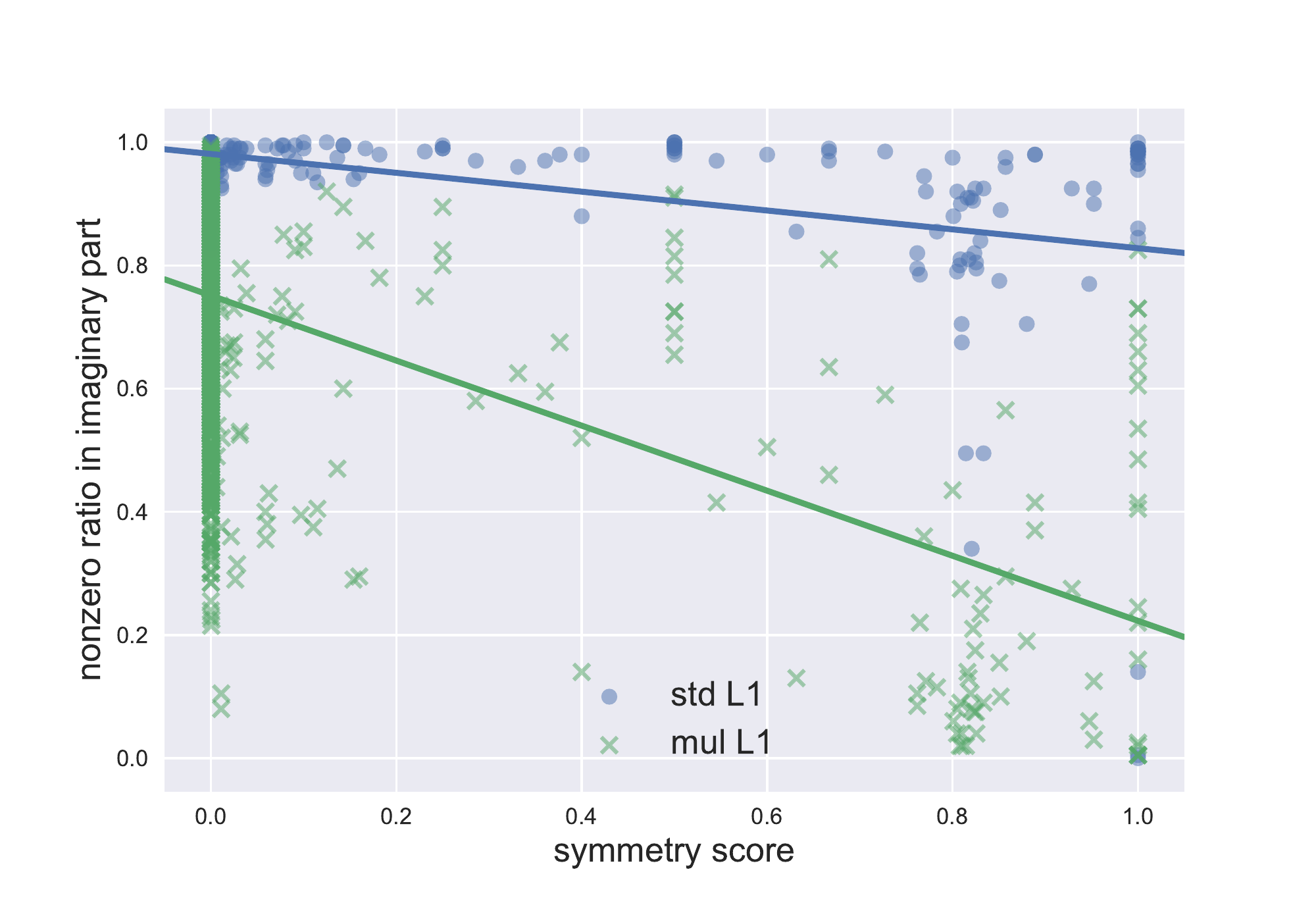}
\subcaption{The imaginary part on the FB15k dataset}
\end{minipage}
\end{tabular}
\caption{The scatter plots showing the degree of sparsity against the symmetry score for each relation.}
\label{fig:scatter}
\end{figure*}

\subsection{Results}

We compared our proposed model (`ComplEx w/ mul L1') with ComplEx and ComplEx with standard L1 regularization (`ComplEx w/ std L1').
Note that our model reduces to ComplEx when $\alpha=0$.

Table~\ref{tab:res} shows the results. %
The results for TransE, DistMult, HolE, and ANALOGY are transcribed from the literature \cite{complex,analogy}.
All the compared models, except for DistMult, can produce a non-symmetric scoring function.

For most of the evaluation metrics,
the multiplicative L1 regularizer (`ComplEx w/ mul L1')
outperformed or was competitive to the best baseline.
Of particular interest, its performance on FB15k
was much better than that of WN18.
Indeed, the accuracy of the original ComplEx is already quite high for WN18.

This difference between two datasets comes from the percentage of infrequent relations in two datasets.
Most of the 1,345 relations in FB15k are infrequent, i.e., each of them has only several dozen triplets in training data.
By contrast, of the 18 relations in WN18, most have more than one thousand training triplets.

Because sparse modeling is, in general, much effective when training data is scarce,
this difference in the proportion of infrequent relations should have contributed to the different performance improvements on the two datasets.
To support this hypothesis, we ran additional experiments with WN18, by reducing the number of training samples to one half. %
The test data remained the same as in the previous experiment.
The results are shown in Table~\ref{tab:res-low}.
The standard and multiplicative L1 regularizers work better than the original ComplEx
and the improvement in each evaluation metric is greater than that observed with all training data.
The proposed method also outperformed the standard L1 regularization consistently.

The reliability of the result in Table~\ref{tab:res} was verified by computing the variance of the filtered MRR scores over 8 trials on FB15k.
For each of the trials, different random choices were used to generate the initial values of the representation vectors and the order of samples to process.
The result, shown in Figure~\ref{fig:boxplot}, confirms that random initial values have little effect on the result;
there is no overlapping MRR range among the compared methods,
and the proposed method (`ComplEx w/ mul L1') consistently outperformed the standard L1 (`ComplEx w/ std L1') and vanilla ComplEx with only L2 regularization.

\subsection{Analysis}

We analyzed how the two L1 regularizers, standard and multiplicative, work for symmetric and antisymmetric
relations on FB15k.
Because of the large number of relations FB15k contains,
manually extracting symmetric/antisymmetric relations is difficult.
To quantify the degree of symmetry of each relation $r$,
we define its ``symmetry score'' by
$ \mathop{\text{sym}}(r) = |\mathcal{T}^{\text{sym}}_r| / |\mathcal{T}_r| $
where
\begin{align*}
  \mathcal{T}_r              & = \{ (s, r, o) \mid (s, r, o) \in \Omega, y_{rso} = +1 \}, \\
  \mathcal{T}^{\text{sym}}_r & = \{ (s, r, o) \mid (s, r, o) \in \Omega, y_{rso} = y_{ros} = +1 \},
\end{align*}
and compute this score for all relations in the FB15k training set.
By definition,
symmetric relation $r$ should give $\text{sym}(r)=1.0$
and
antisymmetric relation $r$ should give $\text{sym}(r)=0.0$.
Thus, this score should indicate the degree of symmetry/antisymmetry of relations.

In Figure~\ref{fig:scatter}, we plot the percentage of non-zero elements in the real and imaginary parts
of its representation vector $\mat{w}_r$ against $\text{sym}(r)$ for each relation $r$.
Panels~(a) and (b) are for $\RE(\mat{w}_r)$ and $\IM(\mat{w}_r)$, respectively.
It is expected that the higher~(resp. lower) the symmetry is, the more imaginary~(resp. real) parts become zero.
With
the multiplicative L1 regularizer (denoted by `mul L1' and green crosses in the figure), 
the density of real parts correlates with symmetry, and the density of imaginary parts inversely correlates with symmetry.
By contrast, correlation is weak or not observable for the standard L1 regularizer (`std L1'; blue circles).
As well as demonstrated on synthetic data,
we conclude that the gains in Table~\ref{tab:res}
come mainly from the more desirable representations
for symmetric/antisymmetric relations learned with the multiplicative L1 regularization.

However, compared to the results on synthetic data shown in Figure~\ref{fig:sparse},
many non-zero values exist in $\RE(\mat{w}_r)$ and $\IM(\mat{w}_r)$
even for completely symmetric/antisymmetric relations.
We suspect that
this is related to the fact that an antisymmetric relations does not imply an antisymmetric truth-value matrix;
even if a relation is antisymmetric, there are many entity pairs $(e_1, e_2)$ such that
neither of $(e_1, r, e_2)$ or $(e_2, r, e_1)$ holds.

\section{Conclusion}

In this paper,
we have presented a new regularizer for ComplEx
to encourage vector representations of relations to be more symmetric or antisymmetric in accordance with data.
In the experiments, the proposed regularizer improved over the original ComplEx (with only L2 regularization)
on FB15k, as well as WN18 with limited training data.

Recently, researchers have been making attempt to leverage background knowledge to improve KBC,
such as by incorporating information of entity types, hierarchy, and relation attributes into the models
\cite{krompass2015type,ECMLPKDD2017}.
Our method does not assume background knowledge but adapt to symmetry/asymmetry of relations in a data-driven manner.
However, learning more diverse knowledge about relations from data should be an interesting future research topic.
We would also like to consider different type of regularization for entity vectors.

As another future research direction, we would like to develop a better strategy for sampling negative triplets
that are suitable for our method.

\section*{Acknowledgments}

We are grateful to the anonymous reviewers for useful comments.
MS thanks partial support by JSPS Kakenhi Grant 15H02749.

\bibliographystyle{aaai}

\begin{thebibliography}{}

\bibitem[\protect\citeauthoryear{Bollacker \bgroup et al\mbox.\egroup
  }{2008}]{freebase}
Bollacker, K.; Evans, C.; Paritosh, P.; Sturge, T.; and Taylor, J.
\newblock 2008.
\newblock Freebase: A collaboratively created graph database for structuring
  human knowledge.
\newblock In {\em Proceedings of the 2008 ACM SIGMOD International Conference
  on Management of Data (SIGMOD '08)},  1247--1250.

\bibitem[\protect\citeauthoryear{Bordes \bgroup et al\mbox.\egroup
  }{2013}]{transe}
Bordes, A.; Usunier, N.; Garcia-Duran, A.; Weston, J.; and Yakhnenko, O.
\newblock 2013.
\newblock Translating embeddings for modeling multi-relational data.
\newblock In {\em Advances in Neural Information Processing Systems 26 (NIPS
  '13)},  2787--2795.

\bibitem[\protect\citeauthoryear{Duchi, Hazan, and Singer}{2011}]{adagrad}
Duchi, J.; Hazan, E.; and Singer, Y.
\newblock 2011.
\newblock Adaptive subgradient methods for online learning and stochastic
  optimization.
\newblock {\em Journal of Machine Learning Research} 12:2121--2159.

\bibitem[\protect\citeauthoryear{Faruqui \bgroup et al\mbox.\egroup
  }{2015}]{overcomplete}
Faruqui, M.; Tsvetkov, Y.; Yogatama, D.; Dyer, C.; and Smith, N.~A.
\newblock 2015.
\newblock Sparse overcomplete word vector representations.
\newblock In {\em Proceedings of the 53rd Annual Meeting of the Association of
  Computational Linguistics (ACL '15)},  1491--^^ef^^bf^^bd1500.

\bibitem[\protect\citeauthoryear{Fellbaum}{1998}]{wordnet}
Fellbaum, C.
\newblock 1998.
\newblock {\em WordNet: An Electronic Lexical Database}.
\newblock MIT Press.

\bibitem[\protect\citeauthoryear{Hayashi and Shimbo}{2017}]{eq}
Hayashi, K., and Shimbo, M.
\newblock 2017.
\newblock On the equivalence of holographic and complex embeddings for link
  prediction.
\newblock In {\em Proceedings of the 55th Annual Meeting of the Association for
  Computational Linguistics (ACL '17)},  554--559.

\bibitem[\protect\citeauthoryear{Kong \bgroup et al\mbox.\egroup
  }{2014}]{exclusivelasso}
Kong, D.; Fujimaki, R.; Liu, J.; Nie, F.; and Ding, C.
\newblock 2014.
\newblock Exclusive feature learning on arbitrary structures via
  $\ell_{1,2}$-norm.
\newblock In {\em Advances in Neural Information Processing Systems 27},
  1655--1663.

\bibitem[\protect\citeauthoryear{Krompa{\ss}, Baier, and
  Tresp}{2015}]{krompass2015type}
Krompa{\ss}, D.; Baier, S.; and Tresp, V.
\newblock 2015.
\newblock Type-constrained representation learning in knowledge graphs.
\newblock In {\em International Semantic Web Conference},  640--655.
\newblock Springer.

\bibitem[\protect\citeauthoryear{Liu, Wu, and Yang}{2017}]{analogy}
Liu, H.; Wu, Y.; and Yang, Y.
\newblock 2017.
\newblock Analogical inference for multi-relational embeddings.
\newblock In {\em Proceedings of the 34th International Conference on Machine
  Learning (ICML '17)},  2168--2178.

\bibitem[\protect\citeauthoryear{Mikolov \bgroup et al\mbox.\egroup
  }{2013}]{word2vec}
Mikolov, T.; Chen, K.; Corrado, G.; and Dean, J.
\newblock 2013.
\newblock Efficient estimation of word representations in vector space.
\newblock {\em arXiv} 1301.3781.

\bibitem[\protect\citeauthoryear{Minervini \bgroup et al\mbox.\egroup
  }{2017}]{ECMLPKDD2017}
Minervini, P.; Costabello, L.; Mu{\~{n}}oz, E.; Nov{\'a}{\v{c}}ek, V.; and
  Vandenbussche, P.-Y.
\newblock 2017.
\newblock Regularizing knowledge graph embeddings via equivalence and inversion
  axioms.
\newblock In {\em Proceedings of the 2017 European Conference on Machine
  Learning and Knowledge Discovery in Databases (ECML PKDD '17)}.

\bibitem[\protect\citeauthoryear{Nickel \bgroup et al\mbox.\egroup
  }{2016}]{survey}
Nickel, M.; Murphy, K.; Tresp, V.; and Gabrilovich, E.
\newblock 2016.
\newblock A review of relational machine learning for knowledge graphs.
\newblock {\em Proceedings of the {IEEE}} 104(1):11--33.

\bibitem[\protect\citeauthoryear{Nickel, Rosasco, and Poggio}{2016}]{hole}
Nickel, M.; Rosasco, L.; and Poggio, T.
\newblock 2016.
\newblock Holographic embeddings of knowledge graphs.
\newblock In {\em Proceedings of the 30th AAAI Conference on Artificial
  Intelligence (AAAI '16)},  1955--1961.

\bibitem[\protect\citeauthoryear{Nickel, Tresp, and Kriegel}{2011}]{rescal}
Nickel, M.; Tresp, V.; and Kriegel, H.-P.
\newblock 2011.
\newblock A three-way model for collective learning on multi-relational data.
\newblock In {\em Proceedings of the 28th International Conference on Machine
  Learning (ICML '11)},  809--816.

\bibitem[\protect\citeauthoryear{Pennington, Socher, and Manning}{2014}]{glove}
Pennington, J.; Socher, R.; and Manning, C.~D.
\newblock 2014.
\newblock Glove: Global vectors for word representation.
\newblock In {\em Proceedings of the 2014 Conference on Empirical Methods in
  Natural Language Processing (EMNLP '14)},  1532--1543.

\bibitem[\protect\citeauthoryear{Suchanek, Kasneci, and Weikum}{2007}]{yago}
Suchanek, F.~M.; Kasneci, G.; and Weikum, G.
\newblock 2007.
\newblock Yago: A core of semantic knowledge.
\newblock In {\em Proceedings of the 16th International Conference on World
  Wide Web (WWW '07)},  697--706.

\bibitem[\protect\citeauthoryear{Sun \bgroup et al\mbox.\egroup
  }{2016}]{Sun:Sparse}
Sun, F.; Guo, J.; Lan, Y.; Xu, J.; and Cheng, X.
\newblock 2016.
\newblock Sparse word embeddings using $\ell_1$ regularized online learning.
\newblock In {\em Proceedings of the 25th International Joint Conference on
  Artificial Intelligence (IJCAI '16)},  2915--2921.

\bibitem[\protect\citeauthoryear{Tibshirani \bgroup et al\mbox.\egroup
  }{2005}]{fused}
Tibshirani, R.; Saunders, M.; Rosset, S.; Zhu, J.; and Knight, K.
\newblock 2005.
\newblock Sparsity and smoothness via the fused lasso.
\newblock {\em Journal of the Royal Statistical Society Series B}
  67(1):91--108.

\bibitem[\protect\citeauthoryear{Tibshirani}{1994}]{lasso}
Tibshirani, R.
\newblock 1994.
\newblock Regression shrinkage and selection via the lasso.
\newblock {\em Journal of the Royal Statistical Society, Series B} 58:267--288.

\bibitem[\protect\citeauthoryear{Trouillon \bgroup et al\mbox.\egroup
  }{2016a}]{unitarydiag}
Trouillon, T.; Dance, C.~R.; Gaussier, {\'E}.; and Bouchard, G.
\newblock 2016a.
\newblock Decomposing real square matrices via unitary diagonalization.
\newblock {\em arXiv} 1605.07103.

\bibitem[\protect\citeauthoryear{Trouillon \bgroup et al\mbox.\egroup
  }{2016b}]{complex}
Trouillon, T.; Welbl, J.; Riedel, S.; Gaussier, E.; and Bouchard, G.
\newblock 2016b.
\newblock Complex embeddings for simple link prediction.
\newblock In {\em Proceedings of the 33rd International Conference on Machine
  Learning (ICML '16)},  2071--2080.

\bibitem[\protect\citeauthoryear{Xiao}{2009}]{rda}
Xiao, L.
\newblock 2009.
\newblock Dual averaging method for regularized stochastic learning and online
  optimization.
\newblock In {\em Advances in Neural Information Processing Systems 22 (NIPS
  '09)}.

\bibitem[\protect\citeauthoryear{Yang \bgroup et al\mbox.\egroup
  }{2015}]{distmult}
Yang, B.; Yih, W.; He, X.; Gao, J.; and Deng, L.
\newblock 2015.
\newblock Embedding entities and relations for learning and inference in
  knowledge bases.
\newblock {\em Proceedings of the 3rd International Conference on Learning
  Representations (ICLR '15)}.

\end{thebibliography}

\end{document}